\documentclass[11pt]{article}
\usepackage[preprint]{acl}

\usepackage{times}
\usepackage{latexsym}
\usepackage{booktabs}
\usepackage{pdflscape}
\usepackage{makecell}
\usepackage{booktabs}  % For better table formatting
\usepackage{siunitx}   % For better number alignment
\usepackage{enumitem}
\usepackage{multirow}
\usepackage{subcaption}
\usepackage{graphicx}
\usepackage{tabularx}
\usepackage{xcolor}
\usepackage{colortbl}
\usepackage{arydshln}
\usepackage{fontawesome5}
\usepackage{float}
\usepackage{tabularx}
\usepackage{amssymb}

\usepackage[T1]{fontenc}
\usepackage{microtype}

\usepackage{inconsolata}
\usepackage{hyperref}

\usepackage{graphicx}

\title{Language-Guided Temporal Token Pruning for Efficient \\VideoLLM Processing}

\author{Yogesh Kumar \\\
  Indian Institute of Technology Jodhpur\\
  \texttt{kumar.204@iitj.ac.in}
  }

\begin{document}
\maketitle

\begin{abstract}
Vision Language Models (VLMs) struggle with long-form videos due to the quadratic complexity of attention mechanisms. We propose Language-Guided Temporal Token Pruning (LGTTP), which leverages temporal cues from queries to adaptively prune video tokens, preserving contextual continuity while reducing computational overhead. Unlike uniform pruning or keyframe selection, LGTTP retains higher token density in temporally relevant segments. Our model-agnostic framework integrates with TimeChat and LLaVA-Video, achieving a 65\% reduction in computation while preserving 97-99\% of the original performance. On QVHighlights, LGTTP improves HIT@1 by +9.5\%, and on Charades-STA, it retains 99.6\% of R@1. It excels on queries with explicit temporal markers and remains effective across general video understanding tasks. The code is available at: \url{https://github.com/yogesh-iitj/LGTTP}.
\end{abstract}

\section{Introduction}
Vision Language Models like TimeChat \cite{ren2024timechat}, LLaVA-Video \cite{zhang2024llavanext-video}, and VideoLLaVA \cite{lin2023video} have demonstrated exceptional capabilities across various video understanding tasks, from highlight detection to temporal grounding and video question answering. However, these models face substantial computational inefficiency when processing long-form videos, as they typically encode every frame into visual tokens and process the entire token sequence for each query. The computational complexity grows quadratically with sequence length due to the attention mechanism, making efficient token management a critical challenge for practical deployment. Real-world video queries often target specific temporal segments, making full sequence processing inefficient.

Current efficiency approaches fall into two categories with significant limitations: (1) vision token pruning methods like PruMerge \cite{shang2024LLaVA-PruMerge} and ToMe \cite{bolya2022tome}, which reduce spatial redundancy within individual frames but fail to capture temporal connections; and (2) keyframe selection methods like KeyVideoLLM \cite{keyvideo} and VideoTree \cite{wang2024videotree}, which disrupt temporal context by completely discarding intermediate frames. These limitations are particularly problematic for temporal understanding tasks that rely on maintaining temporal coherence across frames, such as highlight detection and temporal grounding.  
Moreover, uniform pruning overlooks the dynamic relevance of frames across time, leading to suboptimal retention of critical moments. This motivates the need for query-aware pruning strategies that adaptively preserve temporally salient content.

\begin{figure*}[t!]
    \centering
      \includegraphics[width=0.92\textwidth]{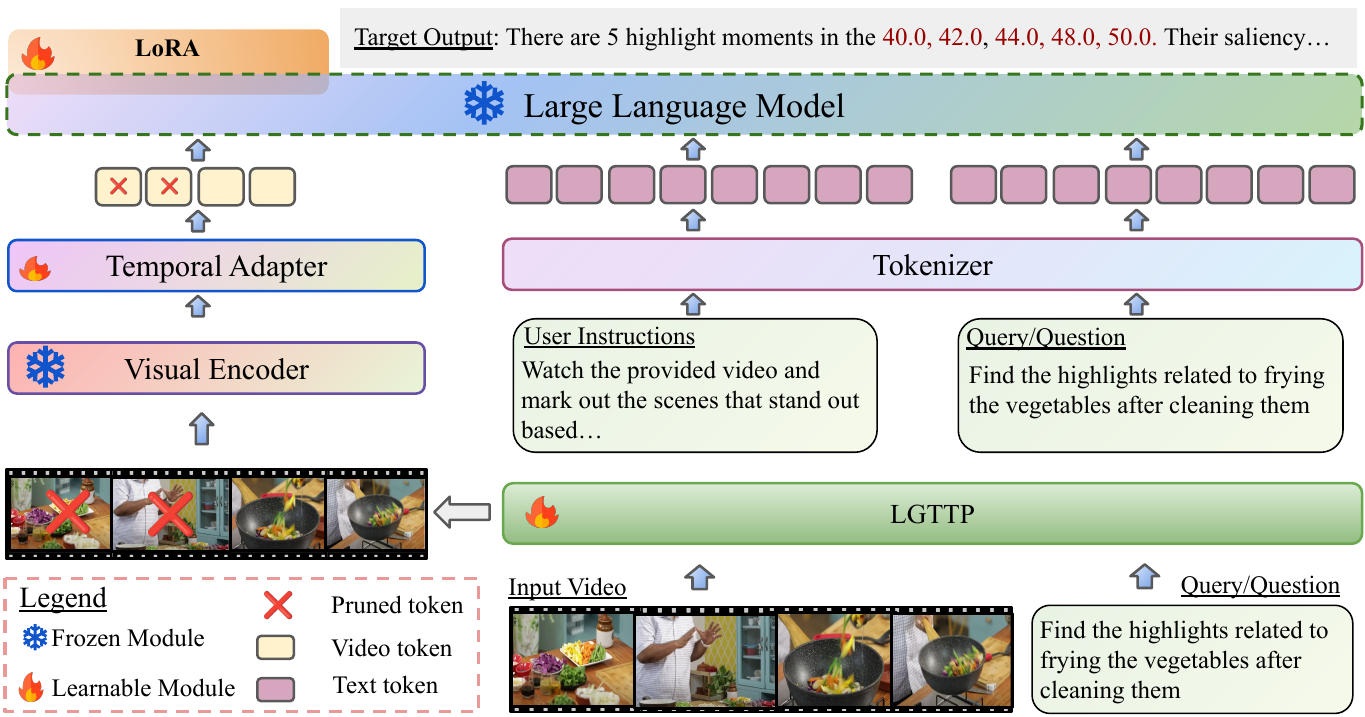}
  \caption{\label{fig:model} Given a target video and a natural language query, LGTTP identifies frames for pruning based on temporal relevance. Representations of the pruned frames are discarded, and only the remaining tokens are forwarded to the Large Language Model for downstream processing.}
\end{figure*}

We propose \underline{L}anguage-\underline{G}uided \underline{T}emporal \underline{T}oken \underline{P}runing (LGTTP), a model-agnostic approach that addresses these limitations by adaptively assigning pruning rates based on temporal cues extracted from queries. LGTTP integrates effectively with both TimeChat and LLaVA-Video architectures, showing particular strength with models that have built-in temporal awareness. By leveraging temporal elements, LGTTP preserves tokens most relevant to temporal queries while reducing computational requirements by 65\% and maintaining 97-99\% of original performance. This enables efficient long-form video processing without compromising temporal coherence or task accuracy.

Specifically, our contributions include: (i) A model-agnostic framework that extracts temporal cues from natural language queries to guide token pruning. (ii) A method to integrate LGTTP with VideoLLM architectures, with particular optimization for temporally-aware models. (iii) Comprehensive evaluation across video understanding benchmarks, demonstrating that LGTTP maintains near-original performance while significantly reducing computational requirements.

\section{\texorpdfstring{\underline{L}anguage-\underline{G}uided \underline{T}emporal \underline{T}oken \underline{P}runing (LGTTP) Framework}{Language-Guided Temporal Token Pruning (LGTTP) Framework}}
\noindent{\textbf{Overview and Motivation.}}
As shown in Figure~\ref{fig:model}, LGTTP addresses a fundamental efficiency challenge in VideoLLMs: not all frames are equally relevant to a given query, particularly for temporal understanding tasks. While existing approaches either prune tokens uniformly or select entire keyframes, LGTTP adaptively preserves the most temporally relevant tokens while maintaining contextual continuity. Our approach consists of the following main components: temporal cue extraction from queries, temporally-aware relevance prediction for different models, and adaptive token pruning based on predicted relevance. By leveraging temporal indicators in queries, we can concentrate computational resources on the most relevant segments of the video.

\noindent{\textbf{Temporal Cue Extraction.}}
Natural language queries often contain rich temporal information that guides our pruning strategy. Given a query $Q$, we extract temporal information through two stages: First, we identify temporal markers (e.g., ``before,'' ``after,'' ``during'') using pattern matching and a fine-tuned classifier, categorizing them into \textit{Precedence}, \textit{Subsequence}, or \textit{Co-occurrence} relationships. Second, we extract reference events and their potential temporal positions, for example, from ``after talking to the coach," we identify ``talking to the coach" as occurring earlier in the video. This temporal knowledge helps prioritize frames likely to contain moments of interest.

\noindent{\textbf{Adaptation to VideoLLM Architectures.}}
LGTTP integrates with various VideoLLM architectures according to their temporal awareness capabilities. Given input sampled frames $f_1, f_2, ..., f_N$, different models generate initial embeddings $E = \{e_1, e_2, ..., e_N\}$ through their respective vision encoders. We then create temporally-adapted embeddings $E' = \{e'_1, e'_2, ..., e'_N\}$ based on the model's capabilities. For timestamp-aware models like TimeChat, we leverage existing timestamp bindings. where the embeddings already incorporate temporal information.
% \begin{equation}
% E' = E = F_{img}(f_1, f_2, ..., f_N, t_1, t_2, ..., t_N)
% \end{equation}

For models using temporal instructions (e.g., LLaVA-Video), we add lightweight temporal position embeddings based on normalized frame positions, where each frame embedding $e_i$ is augmented with $P_{temp}(i/N)$ to create the adapted embeddings $e'_i = e_i + P_{temp}(i/N)$. Here, $P_{temp}(x) = W_p \cdot x + b_p$ is a learned linear function that maps normalized positions to temporal features, where $W_p \in \mathbb{R}^{d \times 1}$ and $b_p \in \mathbb{R}^d$.

For standard VLMs without explicit temporal awareness, we introduce a temporal adapter that projects frame indices into positional embeddings, computing adapted embeddings as $e'_i = e_i + A_{temp}(i)$ for each frame $i$. The temporal adapter function $A_{temp}(i) = \text{scale} \times \text{MLP}(\text{temporal\_embed}(i))$ combines an embedding layer with a two-layer MLP and a learnable scaling factor.

After obtaining temporally-adapted embeddings $E'$, we compute relevance scores by processing the query embedding $e_q = F_{text}(Q)$ and calculating:
\begin{equation}
L_{base} = a \cdot \text{cos\_sim}(E', e_q) + b,
\end{equation}
where $a$ and $b$ are learnable parameters. We then incorporate extracted temporal cues through a weighting mechanism:
\begin{equation}
L_{temp} = L_{base} \odot W_{temp},
\end{equation}
where $W_{temp}$ prioritizes frames based on identified temporal relationships. 

Following, we explain the temporal weights $W_{temp} = \{w_1, w_2, ..., w_N\}$ generation.

\noindent{\textbf{Temporal Weight Generation.}}
The extracted temporal markers and reference events are converted into frame-wise temporal weights $W_{temp} = \{w_1, w_2, ..., w_N\}$ that guide the pruning process. Our weighting strategy mirrors human temporal reasoning, focusing attention on relevant segments while preserving broader context.

For \textit{Precedence} markers (``before,'' ``prior to''), we apply linearly decreasing weights that prioritize earlier frames while preserving later context:
\begin{equation}
w_i = 1.5 - \frac{i-1}{N-1} \quad \text{for } i = 1, 2, ..., N.
\end{equation}
This linear decay ensures that frames closer to the beginning receive higher retention rates (up to $1.5\times$ baseline), while later frames maintain minimum context (down to $0.5\times$ baseline).

For \textit{Subsequence} markers (``after,'' ``following''), we employ the inverse pattern to emphasize later temporal segments:
\begin{equation}
w_i = 0.5 + \frac{i-1}{N-1} \quad \text{for } i = 1, 2, ..., N.
\end{equation}
The linear increase reflects the temporal logic that events ``after'' a reference point are more likely to occur in later portions of the video.

For \textit{Co-occurrence} markers (``during,'' ``while''), we use a Gaussian-like distribution centered on middle frames, motivated by the observation that co-occurring events often happen in the central portion of video segments:
\begin{equation}
w_i = \exp\left(-\lambda \cdot \left|\frac{i-1}{N-1} - 0.5\right|\right),
\end{equation}
where $\lambda$ controls the concentration around the center. This exponential decay ensures sharp focus on central frames while maintaining sufficient context from peripheral frames for temporal coherence.

The weight range of $[0.5, 1.5]$ is chosen to provide meaningful differentiation while preventing excessive token elimination that could disrupt contextual understanding. For queries without explicit temporal markers, we maintain uniform weights $w_i = 1.0$ to avoid introducing bias. When multiple temporal relationships are detected within a single query, we combine their weights through element-wise multiplication followed by normalization to preserve the overall pruning magnitude while capturing the compound temporal focus.

% We use a weight range of $[0.5, 1.5]$ to balance differentiation and contextual preservation. For queries without explicit temporal markers, uniform weights $w_i = 1.0$ are applied to avoid bias. When multiple temporal cues are present, weights are combined via element-wise multiplication and normalized to maintain pruning consistency while capturing compound temporal focus.

% \noindent{\textbf{Temporal Weight Generation.}}
% The extracted temporal markers and reference events are converted into frame-wise temporal weights $W_{temp} = \{w_1, w_2, ..., w_N\}$ that guide the pruning process. For each temporal relationship category, we generate different weighting patterns:

% For \textit{Precedence} markers (``before,'' ``prior to''), we emphasize earlier frames:
% \begin{equation}
% w_i = 1.5 - \frac{i-1}{N-1} \quad \text{for } i = 1, 2, ..., N
% \end{equation}

% For \textit{Subsequence} markers (``after,'' ``following''), we emphasize later frames:
% \begin{equation}
% w_i = 0.5 + \frac{i-1}{N-1} \quad \text{for } i = 1, 2, ..., N
% \end{equation}

% For \textit{Co-occurrence} markers (``during,'' ``while''), we emphasize central frames:
% \begin{equation}
% w_i = \exp\left(-\lambda \cdot \left|\frac{i-1}{N-1} - 0.5\right|\right)
% \end{equation}
% where $\lambda$ controls the concentration around the center.

% For queries without explicit temporal markers, we maintain uniform weights $w_i = 1.0$ for all frames. When multiple temporal relationships are detected, we combine their weights through element-wise multiplication, normalized to maintain the overall magnitude.

\noindent{\textbf{Temporally-Adaptive Token Pruning.}}
The final step converts temporal relevance scores into frame-specific pruning rates:
\begin{equation}
R = (r_1, r_2, ..., r_N) = \alpha N \cdot \text{softmax}(L_{temp}),
\end{equation}
where $\alpha$ controls the overall pruning rate and $N$ is the frame count. This ensures the average pruning rate across frames is approximately $\alpha$, while the distribution varies based on temporal relevance.

Critically, we employ soft selection rather than hard keyframe selection. For each frame, we retain:
\begin{equation}
T_i = \max(T_{min}, \lceil (1 - r_i) \cdot T_{full} \rceil),
\end{equation}
tokens, where $T_{min}$ is the minimum token count (typically 10\% of the original count $T_{full}$). This preserves contextual continuity by maintaining some tokens even from less relevant frames.

\noindent{\textbf{Temporal Marker Classification.}
We develop our temporal marker classifier using a weakly supervised approach without manual annotation. A lexicon of temporal expressions (e.g., ``before,'' ``after,'' ``during,'' ``while,'' ``when'') is constructed from existing NLP resources, and rule-based pattern matching is applied to identify and classify markers in queries from QVHighlights, Charades-STA, and VideoMME.

The resulting dataset trains a 2-layer MLP classifier atop frozen BERT embeddings. To enhance robustness, we apply data augmentation (e.g., synonym replacement, word reordering) and include queries without explicit markers as negative examples. For queries with implicit temporal cues (e.g., ``show the beginning''), we use a predefined vocabulary and relative timeline positions to infer relationships without manual labeling.

\begin{table*}[ht]
\small
\centering
\begin{tabular}{lccccc}
\toprule
\textbf{Method} & \textbf{FLOPs} & \multicolumn{2}{c}{\textbf{Highlight Detection}} & \multicolumn{2}{c}{\textbf{Temporal Grounding}} \\
 & & \multicolumn{2}{c}{\textbf{QVHighlights}} & \multicolumn{2}{c}{\textbf{Charades-STA}} \\
 & & \textbf{mAP} & \textbf{HIT@1} & \textbf{R@1 (IoU=0.5)} & \textbf{R@1 (IoU=0.7)} \\
\midrule
TimeChat (original) & 100 & 21.7 & 37.9 & 46.7 & 23.7 \\
\hline
Random Sampling & 35 & 14.2 & 25.6 & 34.5 & 15.8 \\
ToMe~\cite{bolya2022tome} & 38 & 15.5 & 27.3 & 36.2 & 16.9 \\
PruMerge~\cite{shang2024LLaVA-PruMerge} & 35 & 16.3 & 28.9 & 37.8 & 17.6 \\
KeyVideoLLM~\cite{keyvideo} & 40 & 13.1 & 27.0 & 32.1 & 14.5 \\
KVTP~\cite{liu2025keyframeorientedvisiontokenpruning} & 35 & 19.6 & 34.2 & 42.3 & 21.2 \\
\midrule
\rowcolor{gray!15}LGTTP (Ours) & 35 & \textbf{21.2} & \textbf{43.7} & \textbf{46.5} & \textbf{23.1} \\
\bottomrule
\end{tabular}
\caption{Performance comparison on highlight detection and temporal grounding tasks with TimeChat}
\label{tab:one}
\end{table*}

\section{Experiments}

\subsection{Experimental Setup}
\textbf{Datasets:} We evaluate LGTTP across multiple video understanding benchmarks: QVHighlights \cite{moment_detr} for the highlight detection task with human-written natural language queries; Charades-STA \cite{Gao2017TALLTA} for temporal grounding that evaluates the ability to locate specific activities described in text; VideoMME \cite{fu2024video} for comprehensive video question answering that tests general understanding capabilities across diverse scenarios; and EgoSchema \cite{damonlpsg2023videollama} for egocentric video understanding with narrative-style queries capturing first-person perspectives.

\noindent{\textbf{Metrics:}} We use task-specific evaluation metrics aligned with standard benchmarks. For highlight detection on QVHighlights, we report mAP and HIT@1. For temporal grounding on Charades-STA, we measure R@1 at IoU thresholds of 0.5 and 0.7. For video question answering on VideoMME and egocentric understanding on EgoSchema, we report accuracy on multiple-choice questions. Across all experiments, we measure computational efficiency using FLOPs relative to the original unmodified models.

\noindent{\textbf{Baselines:}} We compare LGTTP against: Original unmodified models. Random token sampling, Uniform pruning methods: PruMerge \cite{shang2024LLaVA-PruMerge} and ToMe \cite{bolya2022tome}, Keyframe selection: KeyVideoLLM \cite{keyvideo}, KVTP \cite{liu2025keyframeorientedvisiontokenpruning}: A recent state-of-the-art approach for video token pruning.

\subsection{Implementation Details}

We implement LGTTP by integrating it with TimeChat and LLaVA-Video pipelines. For base token pruning, we adopt PruMerge due to its effectiveness in preserving important visual information, though our approach is compatible with other pruning methods.

\noindent{\textbf{Training Configuration.}}
The temporal adapter is trained using Xavier uniform initialization for linear layers and $\mathcal{N}(0, 0.02)$ for embeddings. We use AdamW optimizer with learning rate $1 \times 10^{-4}$ and weight decay $0.01$ for $20$ epochs. We set $\lambda = 2.0$. The vision encoder and LLM remain frozen during adapter training. We trained a lightweight Temporal Marker Classifier and Temporal Adapters, keeping the LLMs' weights frozen throughout. 

\noindent{\textbf{Integration Details.}}
The adapter is inserted after the vision encoder but before the linear projection to LLM space. For temporal cue extraction, we use a $2$-layer MLP classifier trained on automatically labeled queries from QVHighlights, Charades-STA, and VideoMME.

\subsection{Results and Discussion}

\noindent{\textbf{Performance on Temporal Tasks.}}
Table~\ref{tab:one} shows LGTTP's performance on highlight detection (QVHighlights) and temporal grounding (Charades-STA) when integrated with TimeChat. Our approach maintains near-original performance (within 0.5-0.8\%) despite reducing computation by 65\%. On QVHighlights, LGTTP achieves 21.2\% mAP and 43.7\% HIT@1, significantly outperforming other efficiency methods, including KVTP (34.2\% HIT@1). For temporal grounding on Charades-STA, LGTTP achieves 46.5\% R@1 at IoU=0.5, matching the original model's performance while using only 35\% of the computation. This demonstrates LGTTP's particular strength in preserving critical temporal relationships when pruning tokens.

\begin{table*}[ht]
\small
\centering
\begin{tabular}{lccccc}
\toprule
\textbf{Method} & \textbf{FLOPs} & \multicolumn{2}{c}{\textbf{Video Question Answering}} & \multicolumn{2}{c}{\textbf{Egocentric Understanding}} \\
 & & \multicolumn{2}{c}{\textbf{VideoMME}} & \multicolumn{2}{c}{\textbf{EgoSchema}} \\
 & & \textbf{7B} & \textbf{72B} & \textbf{7B} & \textbf{72B} \\
\midrule
LLaVA-Video (original) & 100 & 62.6 & 69.5 & 54.2 & 65.8 \\
\midrule
Random Sampling & 35 & 58.3 & 62.4 & 50.7 & 60.5 \\
ToMe~\cite{bolya2022tome} & 38 & 58.9 & 62.9 & 51.5 & 61.2 \\
PruMerge~\cite{shang2024LLaVA-PruMerge} & 35 & 59.8 & 64.5 & 52.5 & 63.1 \\
KeyVideoLLM~\cite{keyvideo} & 40 & 51.3 & 60.5 & 46.8 & 55.2 \\
KVTP~\cite{liu2025keyframeorientedvisiontokenpruning} & 35 & 61.8 & 66.3 & 52.4 & 63.6 \\
\midrule
\rowcolor{gray!15}LGTTP (Ours) & 35 & \textbf{62.0} & \textbf{67.1} & \textbf{53.1} & \textbf{64.0} \\
\bottomrule
\end{tabular}
\caption{Performance on video question answering and egocentric understanding tasks.}
\label{tab:two}
\end{table*}

\noindent{\textbf{Performance on General Video Understanding.}}
Table \ref{tab:two} shows LGTTP's effectiveness when integrated with LLaVA-Video on question answering and egocentric understanding tasks. Notably, LGTTP maintains performance within 0.6-1.1\% of the original models across both 7B and 72B variants. On VideoMME, LGTTP achieves 62.0\% accuracy with the 7B model, which is effectively on par with the unpruned model (62.6\%). This indicates that LGTTP's temporal awareness benefits extend beyond explicit temporal tasks to general video understanding, where maintaining context across frames remains important.

\noindent{\textbf{Query-Dependent Performance.}}
A key observation is that LGTTP's advantage varies with query type. For queries with explicit temporal markers like ``before/after," LGTTP outperforms KVTP by 7.2\% HIT@1 on QVHighlights. For ``during/while" markers, the advantage is 5.8\%, while for queries without explicit temporal markers, it maintains a 2.3\% improvement. This gradation confirms our hypothesis that language-guided pruning is most beneficial when temporal relationships are explicitly expressed, though it provides benefits across all query types.

\noindent{\textbf{Efficiency-Performance Balance.}}
Across all experiments, LGTTP achieves an optimal efficiency-performance balance. On highlight detection, it retains 97.7\% of the original mAP performance, while on temporal grounding, it maintains 99.6\% of R@1 (IoU=0.5) performance. For video QA tasks, it preserves 99.0\% of the accuracy, all while reducing computation by 65\%. This makes LGTTP well-suited for deployment in resource-constrained settings.

\subsection{Ablation Studies}

\noindent{\textbf{Temporal Cue Extraction Impact.}}
Removing the temporal cue extraction component significantly degrades performance, with varying impact across tasks. For highlight detection, we observe a -6.5\% reduction in HIT@1 on QVHighlights, while temporal grounding on Charades-STA drops by -5.1\% in R@1. Even general video QA tasks see a -1.8\% reduction in accuracy, demonstrating that temporal understanding benefits all video tasks to some degree.

\noindent{\textbf{Architectural Integration Strategies.}}
Our comparison of different integration approaches reveals that adaptation strategy matters. The lightweight temporal adapter approach outperforms simple position embedding by 2.1\% on QVHighlights and 1.5\% on VideoMME. This suggests that modeling temporal relationships requires more sophisticated integration, especially for models without built-in temporal awareness.

\noindent{\textbf{Soft vs. Hard Selection Effects.}}
The difference between soft and hard selection is particularly pronounced. When using hard selection (pruning rate = 0 or 1), performance drops significantly across all tasks: -9.3\% HIT@1 on QVHighlights, -7.6\% R@1 on Charades-STA, and -3.5\% accuracy on VideoMME. These results validate our design choice to maintain a minimum token threshold for all frames, preserving contextual continuity that proves crucial for video understanding.

\noindent{\textbf{Pruning Method Integration.}}
LGTTP improves all baseline pruning methods, but the magnitude varies. When combined with PruMerge, we see +14.8\% HIT@1 improvement on QVHighlights and +8.7\% R@1 on Charades-STA. With ToMe, the improvements are +16.4\% and +10.3\% respectively. This demonstrates LGTTP's versatility as a framework that can enhance existing token reduction methods by adding temporal cues.

\section{Conclusion}
In this work, we introduced LGTTP, a query-guided pruning strategy for VideoLLMs that preserves temporal relevance while reducing computational cost. It consistently outperforms existing methods across tasks like highlight detection, temporal grounding, and video QA, enabling efficient long-form video understanding. Future directions include modeling richer temporal relationships and extending LGTTP to emerging multimodal architectures.

\section{Limitations}
Despite LGTTP's effectiveness, it has several limitations: (1) Performance depends on the presence of temporal cues in queries, with reduced benefits for queries lacking explicit temporal markers; (2) Our implementation handles basic temporal relationships but struggles with complex reasoning involving multiple temporal constraints; (3) While the preprocessing overhead is minimal (0.3-0.5\% of total inference time), it requires additional computational steps; and (4) Optimal integration requires architecture-specific adaptations, potentially limiting straightforward deployment across all VideoLLM variants.

\section*{Acknowledgment}
I am grateful to my PhD supervisor, Dr. Anand Mishra, for his invaluable guidance and generous support throughout my PhD. His mentorship and resources played a crucial role in enabling the publication of this work. I also gratefully acknowledge the support of the UGC NET-JRF fellowship, which funded my research.

\bibliography{custom}

\appendix

\section{Appendix}

\section{Related Work}
\subsection{Video Language Models}
Recent advances in video language modeling have produced a variety of architectures with different approaches to temporal understanding. TimeChat \cite{ren2024timechat} incorporates explicit timestamp awareness by binding visual content with corresponding timestamps, enabling accurate temporal localization. LLaVA-Video \cite{zhang2024llavanext-video} uses a temporal instruction preprompt to inform the LLM about the video sampling process. VideoLLaVA \cite{lin2023video, videollama2} employs a learnable temporal embedding within the vision encoder. These temporal modeling approaches represent different trade-offs between architectural complexity and temporal reasoning capability, with timestamp-aware models achieving superior performance on temporal localization tasks but requiring more sophisticated training procedures.

Other approaches like VideoChat \cite{2023videochat}, Vid2Seq \cite{yang2023vid2seqlargescalepretrainingvisual}, and Video-ChatGPT \cite{Maaz2023VideoChatGPT} have also shown promising results in video understanding tasks, but typically process frames independently, making them less optimal for capturing fine-grained temporal relationships. The independent frame processing paradigm, while computationally simpler, fails to leverage the rich temporal dependencies that are crucial for understanding narrative structure and temporal causality in video content.

\subsection{Efficiency Approaches for VLMs}
Several approaches have been proposed to address the computational challenges of vision token processing in VLMs. Vision token pruning methods like PruMerge \cite{shang2024LLaVA-PruMerge}, ToMe \cite{bolya2022tome}, FastV \cite{chen2024image}, Vid-TLDR \cite{choi2024vid}, and DynamicViT \cite{rao2021dynamicvit} focus on reducing token counts based on importance metrics like attention entropy or attention scores. These methods achieve significant computational savings, typically reducing token counts by 50-80\%, but their effectiveness diminishes on long-form videos where temporal relationships become crucial for understanding narrative flow and causal dependencies. However, these methods typically operate on individual frames without considering temporal context, leading to suboptimal pruning decisions when queries require understanding of temporal sequences or relationships between distant frames.

Keyframe selection methods like KeyVideoLLM \cite{keyvideo}, VideoTree \cite{wang2024videotree}, and Koala \cite{koala} identify and retain only the most informative frames, but their hard selection approach disrupts temporal coherence. While these methods can achieve dramatic efficiency gains by discarding entire frames, the binary selection process often eliminates contextual information that is essential for maintaining temporal understanding, particularly for queries that span multiple temporal segments. The challenge becomes more pronounced in videos with sparse but critical temporal cues, where intermediate frames may contain subtle but important contextual information.

KVTP \cite{liu2025keyframeorientedvisiontokenpruning} bridges these approaches by using soft selection to retain some tokens from less relevant frames, but it does not explicitly leverage temporal cues from queries. This represents a significant limitation when processing temporally complex queries that contain explicit temporal markers, as the method cannot adapt its pruning strategy based on the specific temporal requirements expressed in natural language.

\subsection{Video Understanding Tasks and Benchmarks}
The evaluation of VideoLLMs spans multiple specialized benchmarks. For moment retrieval~\cite{univtg,QD-DETR,Taskweave,chapvidmr,kumar2025matr}, QVHighlights \cite{moment_detr} and Charades-STA \cite{Gao2017TALLTA} are commonly used to evaluate a model's ability to locate temporal segments matching natural language queries. General video understanding capabilities are assessed on benchmarks like VideoMME \cite{fu2024video} and EgoSchema \cite{damonlpsg2023videollama}, which include diverse question-answering tasks. Object, visual relationship, and step localization~\cite{coin,Kumar_2023_CVPR,kumar_qdetrv_2024} is evaluated on datasets like NeXT-QA \cite{xiao2021next}, COIN \cite{coin}, and VidVRD~\cite{vidvrd}. These benchmarks present varying degrees of temporal complexity, ranging from simple activity recognition to sophisticated reasoning about temporal relationships, causality, and narrative structure.

\section{Additional Implementation Details}
\noindent{\textbf{Temporal Adapter.}}
For models without explicit temporal awareness, the temporal adapter consists of three components: (1) an embedding layer (\texttt{nn.Embedding}$(128, 768)$) that maps frame indices to positional embeddings, where $128$ represents the maximum number of frames and $768$ matches the vision encoder output dimension; (2) a two-layer MLP with LayerNorm and GELU activation for feature transformation; and (3) a learnable scale parameter (initialized to $0.1$) that controls the contribution strength when adding temporal embeddings to frame embeddings via residual connection. 

\section{Additional Results}
\noindent{\textbf{Cross-Model Analysis.}}
Our cross-architecture experiments reveal that LGTTP provides consistent benefits regardless of the base model, though with varying magnitudes. The largest improvements come when integrated with TimeChat (+9.5\% HIT@1 on QVHighlights compared to KVTP), likely because TimeChat's timestamp-aware architecture provides more precise temporal information for LGTTP to leverage. With LLaVA-Video, gains are more modest but still significant (+0.8\% on VideoMME), showing LGTTP's effectiveness even with standard temporal modeling approaches.

\noindent{\textbf{End-to-End Latency Analysis.}}
Beyond FLOPs reduction, we evaluate practical deployment efficiency by measuring end-to-end latency on NVIDIA A6000 GPU with 128 frames. LGTTP achieves the best latency performance with $1.54\times$ speedup (1.52s vs 2.34s baseline), outperforming KVTP ($1.48\times$), PruMerge ($1.37\times$), and ToMe ($1.36\times$). While the latency improvement (35\%) is lower than the theoretical FLOPS reduction (65\%) due to fixed overheads in data loading, tokenization, and LLM inference, the 54\% throughput increase (0.66 vs 0.43 videos/s) demonstrates significant practical benefits for real-world deployment. This latency advantage is particularly valuable for interactive video applications where response time is critical, as LGTTP's temporal-aware pruning reduces both computational load and memory bandwidth requirements.

\end{document}